\documentclass[preprint,12pt]{elsarticle}

\usepackage{amssymb}
\usepackage{color}
\usepackage{amsmath}

\journal{Neurocomputing}

\begin{document}

\begin{frontmatter}



\title{Reimagining Reality: A Comprehensive Survey of Video Inpainting Techniques}


\author[inst1]{Shreyank N Gowda}

\author[inst1]{Yash Thakre}
\author[inst1]{Shashank Narayana Gowda}
\author[inst2]{Xiaobo Jin}

\affiliation[inst1]{organization={Vidiosyncracy},
            country={India}}
\affiliation[inst2]{organization={Department of Intelligent Science, School of Advanced Technology, Xi'an Jiaotong-Liverpool University},country={China}}

\begin{abstract}
This paper offers a comprehensive analysis of recent advancements in video inpainting techniques, a critical subset of computer vision and artificial intelligence. As a process that restores or fills in missing or corrupted portions of video sequences with plausible content, video inpainting has evolved significantly with the advent of deep learning methodologies. Despite the plethora of existing methods and their swift development, the landscape remains complex, posing challenges to both novices and established researchers. Our study deconstructs major techniques, their underpinning theories, and their effective applications. Moreover, we conduct an exhaustive comparative study, centering on two often-overlooked dimensions: visual quality and computational efficiency. We adopt a human-centric approach to assess visual quality, enlisting a panel of annotators to evaluate the output of different video inpainting techniques. This provides a nuanced qualitative understanding that complements traditional quantitative metrics. Concurrently, we delve into the computational aspects, comparing inference times and memory demands across a standardized hardware setup. This analysis underscores the balance between quality and efficiency—a critical consideration for practical applications where resources may be constrained. By integrating human validation and computational resource comparison, this survey not only clarifies the present landscape of video inpainting techniques but also charts a course for future explorations in this vibrant and evolving field.
\end{abstract}




\end{frontmatter}



\section{Introduction}
In the realm of video understanding, technologies like action recognition~\cite{wang2013action,sun2022human,gowda2021smart} and video object segmentation~\cite{gowda2020alba,yao2020video,hu2018videomatch} have significantly advanced, enhancing our ability to analyze and interpret dynamic scenes. Building upon these developments, video inpainting has emerged as a captivating research field at the intersection of computer vision, image processing, and machine learning. The exponential growth of multimedia content, coupled with the increasing demand for seamless video editing, has underscored the need for automatic methods that can effectively fill in missing or corrupted regions in videos. Video inpainting techniques aim to address this challenge by reconstructing these regions in a visually plausible manner, enabling a wide range of applications such as video restoration~\cite{restore1,lee2021restore}, object removal~\cite{voleti2022mcvd,liu2021fuseformer,zhang2020autoremover}, and content creation~\cite{yun2020videomix,gowda2022learn2augment,gowda2023pixels}.

The fundamental objective of video inpainting is to infer missing or damaged information based on the available surrounding content, often utilizing temporal and spatial cues within the video sequence. This process involves sophisticated algorithms that analyze the context of the video frames and synthesize new content to seamlessly blend with the existing visual information. By leveraging the inherent structure and patterns within the video frames, inpainting algorithms strive to create visually coherent and continuous video streams where the inpainted regions are virtually indistinguishable from the original content. 

This survey paper aims to provide a comprehensive overview of the state-of-the-art video inpainting techniques developed to date. By exploring the advancements in this field, we seek to shed light on the diverse methodologies, innovations, and challenges encountered by researchers. We categorize video inpainting methods based on their underlying principles and highlight their key contributions, enabling readers to grasp the evolution of techniques. To the best of our knowledge, an existing survey for video inpainting does not exist and hence believe this contribution is vital to the field.

We categorize these methods into distinct groups, each characterized by its unique set of approaches 
and underlying principles. Firstly, we discuss Patch-Based Methods, which primarily involve substituting missing areas in a video frame with patches extracted from other parts of the same video. This category includes Exemplar-Based Inpainting~\cite{shih2009exemplar,koochari2010exemplar,lee2019copy,gowda2019colornet}, leveraging best matching patches for filling holes, and Spatio-Temporal Patch Matching~\cite{newson2014video,le2017motion,cai2022devit}, which extends this concept by considering both spatial and temporal elements in patch selection.

Next, we explore Motion-Based Methods. These techniques utilize the motion information inherent in video sequences to guide the inpainting process. Notable approaches within this category include Optical Flow Estimation~\cite{ding2019frame,wu2020future,gao2020flow} and Motion Compensation~\cite{ebdelli2012examplar,march2022euler,liu2021fuseformer}, both striving to maintain temporal coherence in dynamic scenes.

Finally, we delve into the emerging category of Diffusion-Based Methods for Video Inpainting. These methods represent a novel approach, simulating the diffusion of pixel information from adjacent areas to seamlessly restore damaged or missing regions. This section also touches upon hybrid techniques that combine diffusion-based strategies with other methods for enhanced effectiveness~\cite{gu2023flow,wang2023}.

By categorizing these methods in this manner, we not only simplify the complexity inherent in video inpainting but also shed light on the research trajectory and potential advancements within each category. This classification allows for a clearer understanding of the field's current state and the various approaches being explored to address the challenges in video inpainting.

Throughout this survey, we discuss the influence of various video characteristics, including scene complexity, motion dynamics, and temporal coherence, on the performance of video inpainting methods. Additionally, we address the challenges and open problems that need to be tackled in the field. These challenges encompass improving the handling of large and irregular occlusions, preserving fine details and textures, adapting to dynamic scenes with camera motions, and effectively handling diverse video content. By identifying and understanding these challenges, we hope to inspire researchers to develop novel solutions and push the boundaries of video inpainting. In summary, this survey aims to provide researchers and practitioners with a comprehensive understanding of video inpainting techniques, their advancements, and the challenges that lie ahead. By consolidating the knowledge accumulated in this rapidly evolving field, we hope to stimulate further research and innovation.

In addition to reviewing existing literature, we contribute to the field by conducting a practical evaluation of video inpainting methods. We reimplement five prominent papers from different categories of techniques discussed in this survey. These papers are selected based on their impact, representation of different approaches, and availability of code and models. We carefully compare these methods on a selected number of test samples, taking into consideration both the inference speeds and the quality of inpainted outputs. To ensure a comprehensive evaluation, we employ human annotators to validate the quality of the inpainted videos. Human feedback plays a critical role in assessing the perceptual quality and determining the extent to which the inpainted regions blend seamlessly with the surrounding content. By incorporating these human-annotated scores, we aim to provide a more holistic assessment of the inpainting methods, capturing both objective and subjective aspects of the results.

\section{Fundamentals of Video Inpainting}

Video inpainting is a process used in the field of computer vision and graphics to modify a video by filling in missing or undesired parts of the visual data. The term "inpainting" originally comes from the art restoration field, where it refers to the process of reconstructing damaged parts of paintings or photographs. In the context of video, inpainting serves several purposes:

\begin{itemize}
    \item Restoration: This is similar to the original meaning in art restoration. In video, inpainting can be used to restore old or damaged footage, removing artifacts such as scratches, dust, or other types of noise that degrade the quality.
    \item Object Removal: One of the most common uses of video inpainting is to remove unwanted objects or features from a video. This could be anything from an accidental passerby in a shot, to a piece of modern infrastructure in a historical movie scene, or even watermarks.
    \item Special Effects: In filmmaking and video production, inpainting can be used to achieve specific visual effects. For instance, it can be used to create the illusion of magic tricks, disappearing objects, or to manipulate the video in ways that would be difficult or impossible to achieve during filming.
    \item Editing: Video inpainting can also be used in post-production editing to alter the content of a scene without having to reshoot it. This can be more cost-effective and time-efficient.
    \item Privacy and Security: Inpainting can be used to obscure faces, license plates, or other sensitive information in videos to protect privacy or for security reasons.
\end{itemize}

The challenge in video inpainting, as opposed to still image inpainting, is maintaining consistency across frames. The inpainted regions must match the surrounding area in terms of texture, color, and lighting, and must also look natural as the scene moves and changes over time. Advances in AI and machine learning have significantly improved the capabilities and results of video inpainting in recent years.

Video inpainting involves complex techniques that combine the principles of computer vision, image processing, and machine learning. The underlying goal is to create a visually plausible and consistent output that fills in missing or unwanted parts of a video. One of the earlier techniques used for video inpainting is patch-based inpainting~\cite{criminisi2004region,koochari2010exemplar}, which involves copying and pasting patches (small blocks of pixels) from other parts of the video to fill in the missing or undesired areas. The algorithm searches for the most similar patches in terms of texture and color to ensure a visually coherent result, although this technique is more effective for static backgrounds and can struggle with complex, dynamic scenes.

Maintaining temporal consistency is crucial in video inpainting~\cite{patwardhan2005video,patwardhan2007video}. The inpainted areas must not only blend seamlessly with the spatial surroundings in a single frame but also across the temporal dimension. Techniques used for maintaining temporal consistency often involve tracking the movement and changes in the scene over time to ensure that the inpainted regions behave in a physically plausible manner. With advancements in AI, deep learning models, particularly Generative Adversarial Networks (GANs)~\cite{goodfellow2020generative}, have been applied to video inpainting. These models are trained on large datasets and can generate new content that fills in the missing parts of a video, and are particularly effective in handling complex scenes and dynamic objects, offering more realistic and coherent results.

Optical flow~\cite{horn1981determining} algorithms are used to estimate the motion between video frames. This information is crucial for ensuring that the inpainted content aligns~\cite{le2017motion,xu2019deep,zhang2019internal} correctly with the movement in the video, maintaining the illusion of a continuous, unedited scene. Exemplar-based~\cite{shih2009exemplar,koochari2010exemplar} inpainting extends the idea of patch-based inpainting by considering the entire structure and texture of the area surrounding the hole or undesired object. The algorithm finds similar regions in the video and uses them as references to fill in the missing parts, which helps in maintaining both spatial and temporal coherence. In some cases, especially in professional video editing and restoration, inpainting involves a level of manual input or guidance. Users may define the areas to be inpainted, set motion paths, or make adjustments to ensure the desired outcome.

Advanced inpainting systems incorporate elements of scene understanding~\cite{lao2021flow,zeng2020learning}, where the algorithm interprets the context of the scene to make more informed decisions about how to fill in gaps. This can include predicting the trajectory of moving objects or understanding the overall geometry and depth of the scene. Each of these techniques has its strengths and is chosen based on the specific requirements of the video inpainting task, such as the complexity of the scene, the nature of the missing or unwanted content, and the desired level of automation and control. The field is continuously evolving, with ongoing research aiming to improve the realism, efficiency, and versatility of video inpainting methods.

\section{Taxonomy of Video Inpainting Techniques}

Video inpainting is a dynamic field that encompasses various techniques designed to reconstruct missing or corrupted parts in video frames. This section elaborates on a more comprehensive taxonomy of these techniques.

\subsection{Patch-Based Methods}
Patch-based methods involve replacing missing regions in a video frame with patches from other parts of the video.

\subsubsection{Key Ideas}
These methods rely on the assumption that similar patterns or textures exist within different parts of the video, which can be used to fill in missing regions.

\subsubsection{Algorithms and Approaches}
\begin{itemize}
    \item \textit{Exemplar-Based Inpainting}: These methods involve filling the holes by copying the best matching patches from known regions, often using a priority scheme. Shih et al.~\cite{shih2009exemplar} improves video inpainting by extending an image inpainting algorithm with enhanced patch matching and robust tracking, thus reducing "ghost shadows" and ensuring temporal continuity in different motion segments. Koochari et al~\cite{koochari2010exemplar} use large patches to effectively fill missing parts in video frames, separating moving objects from a stationary background and creating a mosaic for better representation of occluded objects. Lee et al.~\cite{lee2019copy} introduce a deep neural network-based "Copy-and-Paste Networks" for video inpainting, using content from reference frames to effectively fill missing regions while maintaining temporal coherency.
    \item \textit{Spatio-Temporal Patch Matching}: These extend exemplar-based inpainting by considering spatial and temporal dimensions for patch matching. Newson et al.~\cite{newson2013towards} enhance PatchMatch~\cite{patchmatch} for spatio-temporal use, significantly speeding up high-resolution video inpainting while addressing over-smoothing issues. Kim et al.~\cite{kim2019deep} use flow sub-networks and mask sub-networks to model PatchMatch. Le et al.~\cite{le2017motion} utilize motion matching ensuring that the inpainted areas align seamlessly with the surrounding video content's motion patterns. Meanwhile, Liu et al.~\cite{liu2021temporal} propose Temporal Adaptive Alignment Network that aligns and adapts temporal features using patch matching. More recently, Cai et al.~\cite{cai2022devit} utilize deformed vision transformers to effectively reconstruct and inpaint missing or corrupted areas in video content by using Deformed Patch-based Homography (DePtH), which improves patch-level feature alignments. 
\end{itemize}

\subsection{Motion-Based Methods}
Motion-based methods utilize the movement information in videos to achieve inpainting, particularly effective in scenes with consistent motion.

\subsubsection{Key Ideas}
These techniques use motion estimation to guide the inpainting process, ensuring temporal coherence and consistency in dynamic scenes.

\subsubsection{Algorithms and Approaches}
\begin{itemize}
    \item \textit{Optical Flow Estimation}: Zhang et al.~\cite{zhang2019internal} propose repairing or reconstructing missing or damaged areas in video sequences using internal learning mechanisms, leveraging the video's own data and optical flow for inpainting. Work by Ding et al.~\cite{ding2019frame} incorporates ConvLSTM and optical flow for efficient, real-time processing of large and arbitrary length videos. Wu et al.~\cite{wu2020future} develop an approach for predicting future video frames based on a sequence of past continuous video frames using optical flow. More recently, Li et al.~\cite{li2022towards} use a comprehensive framework for video inpainting, utilizing three key modules: flow completion, feature propagation, and content hallucination, to facilitate an efficient and effective inpainting process. Kang et al.~\cite{kang2022error} propose a framework that enhances flow-based video inpainting methods by introducing a newly designed flow completion module and an error compensation network that leverages an error guidance map, aiming to offset the weaknesses of traditional flow-based methods. Zhang et al.~\cite{zhang2023exploiting,zhang2022flow} enhance video inpainting using optical flow guidance in a transformer-based framework. Gao et al.~\cite{gao2020flow} propose video completion  that combines optical flow and edge information.
    
    \item \textit{Motion Compensation}: These methods involve aligning frames based on estimated motion to maintain consistency. Kim et al.~\cite{kim2019recurrent} propose an approach extending deep learning-based image inpainting methods to the video domain which involves an additional time dimension. Ebdelli et al.~\cite{ebdelli2012examplar} present an algorithm for video inpainting that estimates unknown pixels as a linear combination of the closest patches using motion-compensated neighbor embedding. March et al.~\cite{march2022euler} delve into the mathematical underpinnings of motion compensated inpainting in video processing, focusing on the application of Euler equations and analysis of minimizers in this context. Zhang et al.~\cite{zhang2023pfta} introduce PFTA-Net, a novel network that incorporates a progressive feature alignment module with sub-alignments and a progressive refinement scheme for more accurate motion compensation in video inpainting. Liu et al. propose Fuseformer~\cite{liu2021fuseformer}, a transformer-based architecture for video inpainting that enhances the quality of reconstructed video by integrating fine-grained details across frames.
\end{itemize}

\subsection{Diffusion-Based Methods for Video Inpainting}

Diffusion-based methods represent a novel approach in the field of video inpainting, utilizing advanced techniques to enhance the restoration of missing or corrupted parts in video sequences.

\subsubsection{Key Ideas}
Diffusion-based video inpainting leverages the concept of iteratively refining the inpainted area by simulating the diffusion of pixel information from the surrounding areas. This process gradually introduces information into the damaged or missing regions, ensuring a natural and seamless restoration.

\subsubsection{Algorithms and Approaches}
\begin{itemize}
    \item \textit{Extending Image-Based Diffusion to Video}: Recent advancements involve deep learning models that simulate diffusion processes. These models are trained to predict the natural flow of pixel information, making them highly effective for complex video sequences with varying textures and motion. The "Pix2video" work completed by Ceylan et al.~\cite{ceylan2023pix2video} introduces a training-free, text-guided video editing method using image diffusion models that edit an anchor frame and then propagate changes to subsequent frames. Hoppe et al.~\cite{hoppe2022diffusion} introduce the Random-Mask Video Diffusion (RaMViD) method, which extends image diffusion models to video using 3D convolutions and a new conditioning technique, enabling video prediction, infilling, and upsampling. Cherel et al.~\cite{cherel2023infusion} present a technique using internal diffusion processes for video inpainting, focusing on improving the quality and consistency of inpainted video content. Voleti et al~\cite{voleti2022mcvd} introduce a versatile video diffusion model that excels in video prediction, generation, and interpolation by leveraging a masked conditional approach.
    \item \textit{Hybrid Approaches}: Combining diffusion-based techniques with other methods, such as optical flow or patch-based inpainting, allows for more robust solutions that can handle a variety of inpainting challenges in videos. Gu et al.~\cite{gu2023flow} introduce the Flow-Guided Diffusion model for Video Inpainting (FGDVI), which enhances temporal consistency and inpainting quality in videos by reusing an existing image generation diffusion model. Wang et al.~\cite{wang2023} present a method that combines local and nonlocal flow guidance to improve the quality and consistency of video inpainting.
\end{itemize}

Whilst a lot of methods can be classified broadly within these groups, there are approaches which would fit loosely and hence we believe such a style of classification of methods would be easier to summarize and understand. This also gives us an idea on the direction of research among these methods and the potential to improve them.

\section{Challenges in Video Inpainting}

 We identify and discuss key factors such as scene complexity, motion dynamics, and temporal coherence, which significantly impact the performance and applicability of inpainting techniques.

\subsection{Scene Complexity}
Scene complexity in videos, characterized by varying textures, colors, and patterns, poses a substantial challenge for inpainting algorithms. The ability to accurately reconstruct complex scenes while maintaining visual realism is critical. As outlined in multiple previous works~\cite{lao2021flow,newson2013towards,newson2014video}, the complexity of a scene directly influences the choice and performance of inpainting methods, necessitating more advanced techniques for detailed and intricate scenes.

\subsection{Motion Dynamics and Temporal Coherence}
The dynamics of motion within a video significantly affect the inpainting process. Effective inpainting must account for and replicate the natural motion patterns to ensure seamless integration of the inpainted region into the video. This aspect is particularly challenging in videos with rapid or unpredictable motion. We already speak about methods with a specific focus on motion compensation and optical flow prediction. This emphasizes the importance of incorporating motion understanding in video inpainting, especially in high-motion scenes. Maintaining temporal coherence, the consistency of visual elements over time, is essential for the believability and quality of inpainted videos. Temporal incoherencies can lead to noticeable artifacts in video inpainting, thus undermining the overall quality. Advanced methods strive to ensure that inpainted regions remain consistent and coherent throughout the video duration.

\subsection{Challenges and Open Problems}
We also address several pressing challenges and open problems in the field of video inpainting:

\begin{itemize}
    \item \textbf{Handling Large and Irregular Occlusions}: Dealing with large and irregularly shaped occlusions~\cite{ke2021,patwardhan2005video,wang2007video} remains a significant challenge, requiring innovative approaches to ensure effective and realistic inpainting.
    \item \textbf{Preserving Fine Details and Textures}: The preservation of fine details and textures, crucial for realism, is a complex task, especially in highly textured areas. Methods for enhancing texture fidelity in video inpainting has been a strong research area~\cite{wang2020structure,zhang2019internal,propainter}.
    \item \textbf{Adapting to Dynamic Scenes with Camera Motions}: Dynamic scenes, especially those with camera motions, pose unique challenges. As explored previously~\cite{patwardhan2007video,granados2012background,le2017motion}, adapting inpainting methods to accommodate camera movements is essential for realistic reconstructions.
    \item \textbf{Effectively Handling Diverse Video Content}: The diversity of video content, ranging from simple to complex scenes, requires versatile and robust inpainting solutions. 
\end{itemize}

 By consolidating the knowledge accumulated in this rapidly evolving field, we hope to stimulate further research and innovation. Understanding these challenges and the influence of various video characteristics is crucial for the development of more effective and versatile video inpainting methods.

\section{Evaluation Metrics}

In the realm of video inpainting, a key challenge is to evaluate the effectiveness of various algorithms in reconstructing video sequences. The quality of video inpainting is not just about replacing missing or damaged areas; it's about doing so in a way that the viewer perceives the video as uninterrupted and authentic. To quantify this, several evaluation metrics are employed, each focusing on different aspects of video quality and perceptual integrity.

\textbf{Peak Signal-to-Noise Ratio (PSNR)}: PSNR is a commonly used metric for measuring the quality of reconstruction in video and image processing. It calculates the ratio between the maximum possible power of a signal and the power of distorting noise that affects the fidelity of its representation. In video inpainting, PSNR is used to compare the inpainted video against a ground truth or original video to assess the quality of reconstruction.

The formula for PSNR is given by:
\[
PSNR = 10 \cdot \log_{10}\left(\frac{{\text{MAX}^2}}{{\text{MSE}}}\right)
\]
where $\text{MAX}$ is the maximum possible pixel value, and $\text{MSE}$ is the Mean Squared Error between the original and inpainted videos.

\textbf{Structural Similarity Index (SSIM)}: SSIM is a perception-based metric that measures the similarity between two images or videos. It considers changes in texture, luminance, and contrast, providing a more comprehensive assessment than PSNR. In video inpainting, SSIM is utilized to evaluate how well the inpainted areas blend with the surrounding context in terms of perceived changes in structural information.

The formula for SSIM is a combination of luminance, contrast, and structure terms:
\[
SSIM(x, y) = \frac{{(2\mu_x\mu_y + C_1)(2\sigma_{xy} + C_2)}}{{(\mu_x^2 + \mu_y^2 + C_1)(\sigma_x^2 + \sigma_y^2 + C_2)}}
\]
where $\mu_x$, $\mu_y$ are the local means, $\sigma_x$, $\sigma_y$ are the local standard deviations, $\sigma_{xy}$ is the local cross-covariance, and $C_1$ and $C_2$ are constants to stabilize the division.

\textbf{Frechet Inception Distance (FID):} FID is a metric that compares the distribution of generated images to real images in an embedding space. It evaluates both the visual quality of individual frames (e.g., clarity, color accuracy) and the temporal consistency across frames in the video. This is crucial in video inpainting, where maintaining a seamless and consistent flow from frame to frame is as important as the quality of individual frames.

The formula for FID is given by:
\[
FID = \|\mu_{\text{real}} - \mu_{\text{generated}}\|^2 + \text{Tr}(\Sigma_{\text{real}} + \Sigma_{\text{generated}} - 2(\Sigma_{\text{real}}\Sigma_{\text{generated}})^{1/2})
\]
where $\mu_{\text{real}}$ and $\Sigma_{\text{real}}$ are the mean and covariance matrix of feature vectors from the real video, and $\mu_{\text{generated}}$ and $\Sigma_{\text{generated}}$ are the mean and covariance matrix of feature vectors from the generated video.

\textbf{Video Frechet Inception Distance (VFID):} VFID adapts the concept of FID for video. It extends the evaluation to assess not only the quality of individual frames but also the temporal consistency between frames in the generated video compared to the real video.

The formula for VFID is an extension of FID, considering both the frame-wise features and the temporal dynamics:
\[
VFID = FID_{\text{frames}} + \alpha \cdot FID_{\text{temporal}}
\]
where $FID_{\text{frames}}$ is the FID calculated for individual frames, $FID_{\text{temporal}}$ is the FID calculated for temporal dynamics, and $\alpha$ is a weighting factor to balance their contributions.

\textbf{Learned Perceptual Image Patch Similarity (LPIPS):} Learned Perceptual Image Patch Similarity (LPIPS), initially developed for assessing the perceptual similarity between images, has been adapted and extended for use in evaluating videos. This metric is particularly valuable in fields like video generation, video compression, and video editing, where it's important to assess how perceptually similar a generated or altered video is to a reference video. LPIPS measures the distance in VGGNet feature space as a “perceptual loss” for image regression problems.

In video inpainting evaluation, each metric offers a distinct perspective on the quality and effectiveness of video restoration. Peak Signal-to-Noise Ratio (PSNR) is a basic, technical measure that compares videos pixel-by-pixel, focusing on the accuracy of signal reconstruction but not on human visual perception. In contrast, the Structural Similarity Index (SSIM) aligns more closely with human vision by assessing the luminance, contrast, and structural similarity between the original and inpainted videos, making it better suited for evaluating the perceptual integrity of inpainted areas. Video Frechet Inception Distance (VFID) takes a broader approach by evaluating both the visual quality of individual frames and the temporal consistency across a video sequence, thus ensuring a seamless flow in the inpainted video. Lastly, Learned Perceptual Image Patch Similarity (LPIPS) extends the perceptual evaluation to a more nuanced level, focusing on the perceptual similarity of video patches, which is particularly relevant for tasks like video generation and editing where the subtle nuances of human perception play a significant role. Each metric, therefore, caters to different aspects of video quality, from basic signal accuracy to complex perceptual fidelity.

\subsection{Challenges and Limitations in Evaluating Video Inpainting Techniques}

\textbf{Subjective Nature of Quality Assessment:} The assessment of video quality is often subjective, depending on human perception. Metrics like PSNR and SSIM may not fully capture the aesthetic and contextual quality perceived by viewers. This subjectivity can lead to discrepancies between quantitative metric scores and actual perceived quality.

\textbf{Complexity of Dynamic Content:} Video inpainting involves dealing with dynamic content and complex backgrounds, which are difficult to quantify with standard metrics. Metrics may not adequately account for the challenges in handling diverse and complex video content, leading to an incomplete evaluation.

\textbf{Over-Reliance on Quantitative Metrics:} There is often an over-reliance on quantitative metrics like PSNR and SSIM, which might not fully encompass the intricacies of video inpainting. This reliance can overshadow other important aspects such as user experience, realism, and context-awareness.

We show these limitations with varying user scores for different models in Section~\ref{sec:exp} and talk about how subjective video inpainting is and the need for a different metric to properly evaluate models.

\section{Datasets}

\begin{table}[ht]
\centering
\begin{tabular}{|l|l|l|l|l|}
\hline
\textbf{Dataset Name} & \textbf{Focus} & \textbf{Content Description} & \textbf{Resolution} \\ \hline
DAVIS & VOS & Diverse objects & Full HD \\ \hline
YouTube-VOS & VOS & Diverse objects & Various  \\ \hline
DEVIL & Inpainting & Camera Motion & Various \\ \hline
\end{tabular}
\caption{Summary of Datasets for Video Inpainting Research}
\label{tab:datasets}
\end{table}

YoutubeVOS~\cite{ytvos} and DAVIS~\cite{davis} are two extensively utilized datasets in the field of video inpainting, each originally designed for specific tasks unrelated to video inpainting. YoutubeVOS, primarily developed for Video Object Segmentation, offers a diverse range of videos featuring various objects and scenarios, making it a valuable resource for understanding object motion and consistency in videos. DAVIS, on the other hand, was designed for video object segmentation and tracking, providing high-quality, full-resolution video sequences with annotated objects, which are crucial for training models to understand object movement and boundary information.

Despite their initial purposes, both datasets have been adapted for use in video inpainting tasks. This adaptation is primarily due to their comprehensive and varied content, which helps in training inpainting models to handle different scenarios and object movements. However, the lack of datasets specifically designed for video inpainting led to the introduction of the DEVIL~\cite{devil} dataset. This dataset is tailored specifically for video inpainting, encompassing a wider range of challenges specifically encountered in this task, such as dealing with different types of occlusions and complex object interactions.

The absence of a strong, dedicated benchmark in video inpainting has historically hindered the field's progress. This is because generic datasets like YoutubeVOS and DAVIS, while useful, do not fully address the unique challenges of video inpainting, such as temporal consistency and complex dynamic backgrounds. The creation of DEVIL marks a significant step towards overcoming these challenges, providing a more focused and relevant benchmark for advancing the state of the art in video inpainting. These datasets are summarized in Table~\ref{tab:datasets}.

\section{Quality Evaluation and Comparison}
\label{sec:exp}

We run all experiments using a Tesla T4 GPU. We evaluate the following models: Flow-Guided Transformer for Video Inpainting (FGT)~\cite{zhang2022flow}, Flow-edge Guided Video Completion (FGVC)~\cite{gao2020flow}, End-to-End Framework for Flow-Guided Video Inpainting (E2FGVI)~\cite{li2022towards}, FuseFormer~\cite{liu2021fuseformer} and Copy-Paste networks~\cite{lee2019copy}. These methods range from 2019 to more recent methods and showcase how the outputs of these models have evolved over time. First, we compare quantitatively these methods in Table~\ref{tab:exp}. In particular, we look at the inference details of these methods with a view to them being deployed. From these selected models, E2FGVI seems the most realistically deployable model accounting for speed and memory usage. We also report average scores by asking 20 human annotators to give a score on 1 to 10 for each model on a set of videos. The average score is reported under User Rating in the Table.

\begin{table}[]
\centering
\begin{tabular}{|c|c|c|c|c|}
\hline
Model      & Input & Time & Memory & User Rating \\
\hline
FGT~\cite{zhang2022flow} [ECCV'22]        & 80f   & 241s & 2.6 GB & 7.2 \\
FGVC~\cite{gao2020flow}  [ECCV'20]       & 15f   & 50s  & 1.4GB & 6.5 \\
E2FGVI~\cite{li2022towards} [CVPR'22]     & 70f   & 18s  & 8.0GB & 6.8 \\
FuseFormer~\cite{liu2021fuseformer} [ICCV'21] & 50f   & 39s  & 10.5GB & 6.9\\
CopyPaste~\cite{lee2019copy} [ICCV'19]  & 70f   & 44s  & 2.6GB & 6.7\\
\hline
\end{tabular}
\caption{Comparing inference hardware requirements of selected models and the number of frames needed as input to the model.}
\label{tab:exp}
\end{table}

\begin{figure}
    \centering
    \includegraphics[width=0.99\textwidth]{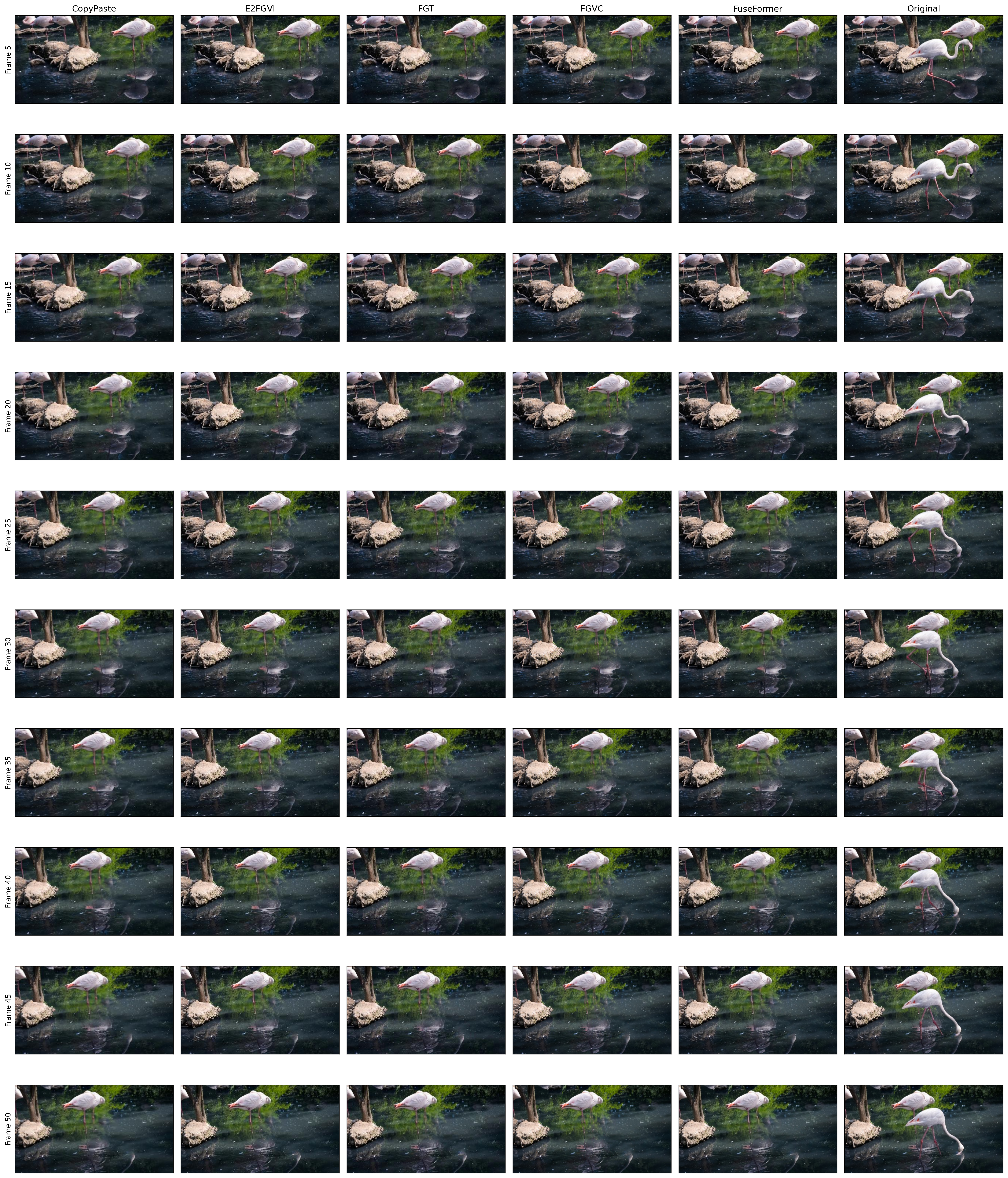}
    \caption{Comparison of different video inpainting methods on a flamingo video. Different frames are taken from the output and shown along with the original frame (ground truth).}
    \label{fig:flamingo}
\end{figure}

From a qualitative perspective, we evaluate on samples from both the evaluation set of datasets the model has been trained on and in-the-wild videos taken from public sources~\footnote{https://mixkit.co/free-stock-video/}. First, we look at a video containing flamingoes and methods that remove one of them. This can be seen in Figure~\ref{fig:flamingo}. 

\begin{figure}
    \centering
    \includegraphics[width=0.99\textwidth]{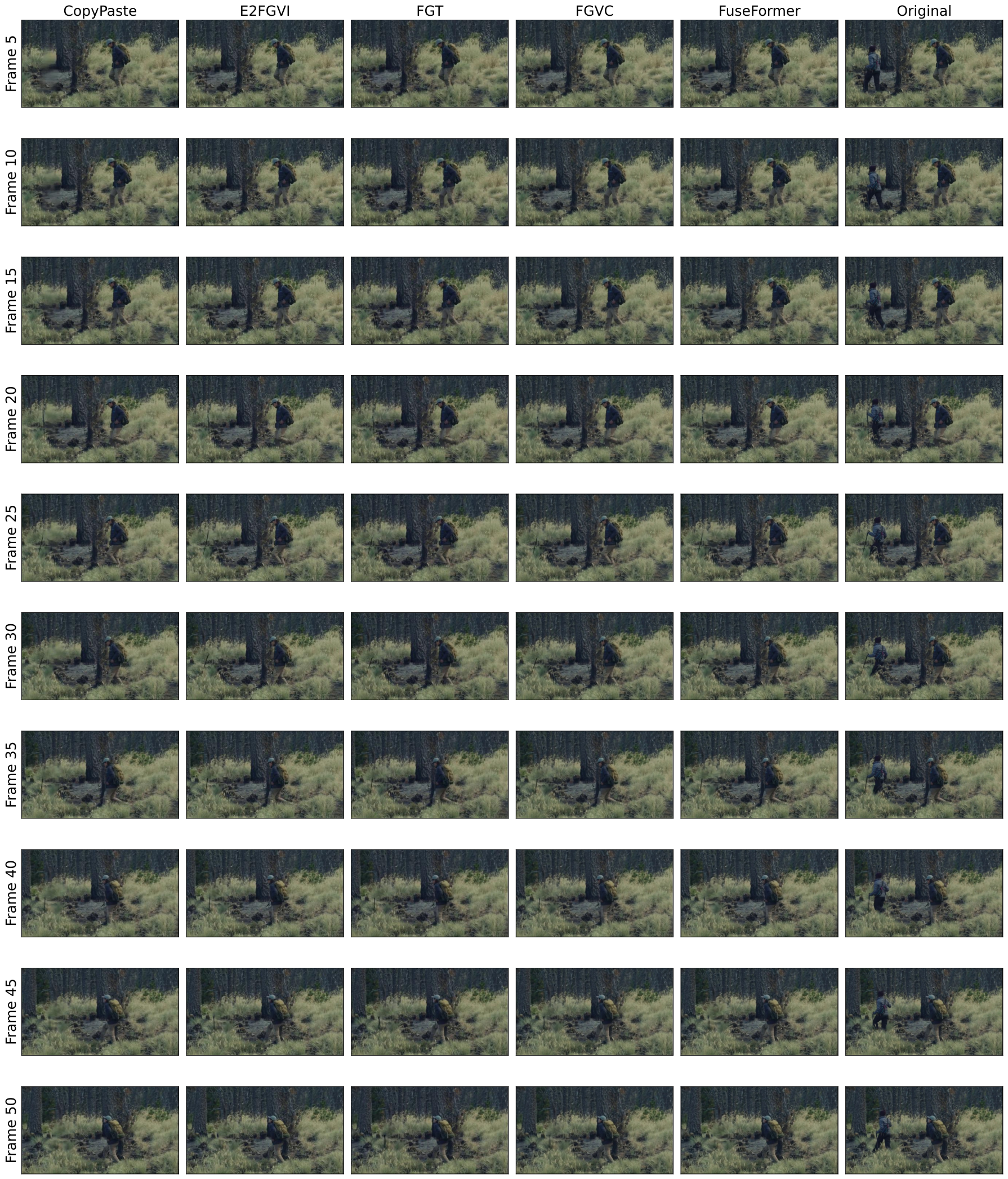}
    \caption{Comparison of different video inpainting methods on a hiking video. Different frames are taken from the output and shown along with the original frame (ground truth).}
    \label{fig:hikers}
\end{figure}

Next, we use a video of people hiking and plot the frame wise results of removal in Figure~\ref{fig:hikers}.

\begin{figure}
    \centering
    \includegraphics[width=0.99\textwidth]{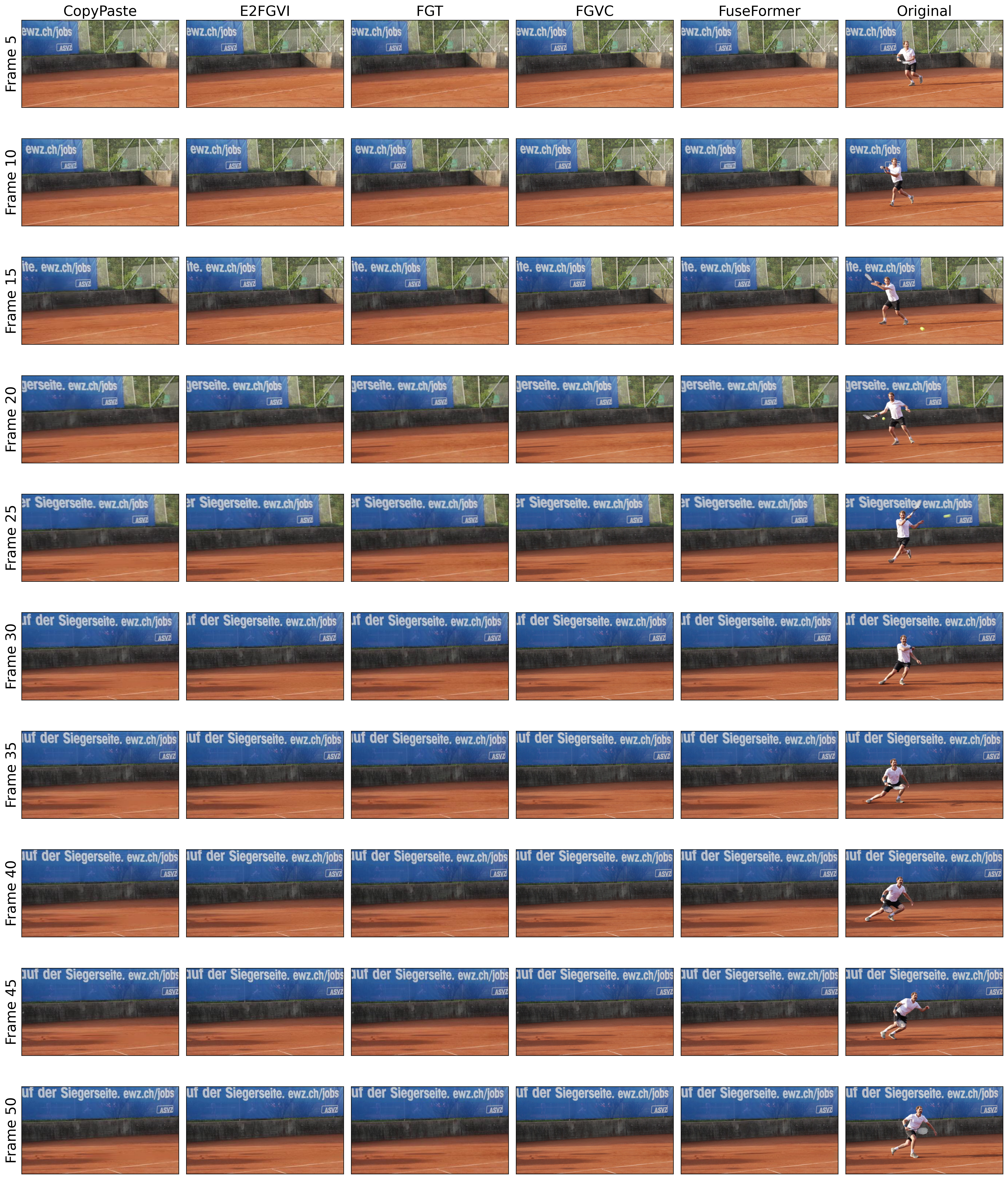}
    \caption{Comparison of different video inpainting methods on a tennis video. Different frames are taken from the output and shown along with the original frame (ground truth).}
    \label{fig:tennis}
\end{figure}

Next, we use a video of person playing tennis and plot the frame wise results of removal in Figure~\ref{fig:tennis}.

\section{Applications and Future Directions}

\subsection{Practical Applications of Video Inpainting}

\begin{itemize}
    \item \textbf{Video Editing:} Video inpainting is a pivotal tool in professional video editing, used for removing unwanted elements like microphones, wires, or unintentional background objects. This technology has found applications in various sectors including film production, news broadcasting, and content creation for digital platforms. It enhances the visual appeal and professionalism of videos by enabling editors to maintain continuity and visual coherence in their narratives.
    
    \item \textbf{Restoration:} Inpainting plays a critical role in the restoration of old or damaged footage. This includes repairing scratches, dust, and other defects in archival footage. Notable projects in film restoration have demonstrated how inpainting helps preserve historical and cultural content, providing clearer and more accessible versions of historically significant films and documentaries.
    
    \item \textbf{Visual Effects:} In the realm of visual effects, inpainting serves as a tool for creating seamless composites and backgrounds. It facilitates the blending of various elements in a scene, aiding in storytelling and the creation of fantastical environments. The use of inpainting in major film productions for generating realistic and visually stunning effects has become increasingly prevalent.
\end{itemize}

\subsection{Current Trends and Emerging Research Directions}


\begin{itemize}
    \item \textbf{Diffusion Models:} Recent trends~\cite{ceylan2023pix2video,voleti2022mcvd} show the integration of diffusion models in video inpainting. These models, based on advanced machine learning algorithms, predict and recreate complex scenes with remarkable accuracy, surpassing traditional techniques in both quality and efficiency.
    
    \item \textbf{Real-Time Inpainting:} The pursuit of real-time video inpainting~\cite{realtime1,realtime2,realtime3} aims to enable live modifications of video streams. This advancement is particularly significant for live broadcasting and virtual reality applications, where the ability to alter or enhance video content instantaneously is invaluable.
    
    \item \textbf{Increasing Automation:} The trend~\cite{ceylan2023pix2video,zhang2020autoremover} towards more automated inpainting processes is making this technology increasingly accessible. Automation reduces the need for manual intervention, thereby streamlining the video editing process and making it more efficient for users with varying levels of expertise.
\end{itemize}

\subsection{Open Challenges and Potential Future Developments}

\begin{itemize}
    \item \textbf{Handling Complex Dynamics:} A significant challenge in video inpainting is dealing with intricate and rapid movements within videos. Future research may focus on developing more sophisticated algorithms that can understand and replicate complex dynamics, thus enhancing the realism and fluidity of inpainted content.
    
    \item \textbf{Enhanced Resolution and Quality:} As video standards continue to evolve towards higher resolutions like 4K and 8K, there is a growing need for inpainting techniques that can maintain detail and clarity at these levels. This development is crucial for ensuring that inpainting technology remains relevant and effective in the face of rapidly advancing display technologies.
    
    \item \textbf{Interactivity and User Control:} Future developments in video inpainting might include more interactive tools, providing users with greater control over the inpainting process. This would allow for more personalized and creative applications, catering to a wider range of use cases and artistic visions.
\end{itemize}

\section{Conclusion}
In conclusion, this paper has provided a thorough examination of the current state of video inpainting techniques, a vital niche in the realms of computer vision and artificial intelligence. Video inpainting, which involves the restoration or completion of missing or corrupted segments in video sequences, has significantly advanced through the integration of deep learning methodologies. Our comprehensive study breaks down key techniques, their foundational theories, and their real-world applications, illuminating a landscape that, despite its rapid evolution, remains complex and challenging.
Central to our analysis was an exhaustive comparative study focusing on visual quality and computational efficiency. By employing a human-centric approach for assessing visual quality, involving a panel of annotators, we gained valuable qualitative insights that enhance traditional quantitative assessments. This human validation aspect brought a unique perspective to our evaluation of various video inpainting methods. In tandem, we conducted a rigorous examination of the computational demands of these techniques, comparing inference times and memory requirements across a uniform hardware setup. Our findings highlight the crucial balance between quality and efficiency, especially pertinent in practical scenarios where resource limitations are a common constraint. This paper not only clarifies the intricate landscape of video inpainting techniques but also sets a direction for future research in this dynamic and rapidly evolving field. By merging human-centric evaluations with computational resource analysis, our study contributes to a deeper understanding of video inpainting and lays the groundwork for future innovations that are both high in quality and efficient in execution.

\section{Acknowledgements}
This work was partially supported by “Qing Lan Project” in Jiangsu universities, NSFC under No. U1804159 and RDF with No. RDF-22-01-020.

 \bibliographystyle{elsarticle-num} 
 \bibliography{cas-refs}





\end{document}